\documentclass[11pt]{article}

\usepackage[final]{acl}

\usepackage{times}
\usepackage{latexsym}

\usepackage[T1]{fontenc}

\usepackage[utf8]{inputenc}

\usepackage{microtype}

\usepackage{inconsolata}

\usepackage{graphicx}

\usepackage{booktabs}

\usepackage{todonotes}

\usepackage{multirow}

\title{CrisiLangs: A Multilingual Resource for Crisis Warning Generation}

\author{
 \textbf{Giacomo Gonella\textsuperscript{1, 2}},
 \textbf{Stefano Menini\textsuperscript{1}},
 \textbf{Marco Guerini\textsuperscript{1}}
\\
\textsuperscript{1}Fondazione Bruno Kessler, Italy
\\
\textsuperscript{2}University of Trento, Italy
\\
 \texttt{\{ggonella, menini, guerini\}@fbk.eu}
}

\begin{document}
\maketitle
\begin{abstract}
Communicating with the public during an emergency entails issuing warning messages promptly. In Europe, it often requires producing them in more than one language. Recent work has shown that language models can be trained to generate warning messages as a crisis unfolds, but the available data covers only English. To fill this gap, we present \textit{CrisiLangs}, a multilingual extension of \textit{CrisiText} (an English dataset of crisis events paired with warning messages written according to expert communication guidelines) to Italian, Bulgarian, German, and Greek. Expert native speakers translated the warning messages of 1,000 events drawn from 411 scenarios, for a total of 5,000 event-message pairs across five languages. We test the resource to fine-tune a multilingual model on all five languages jointly, with and without the emergency guidelines in the prompt, and compare it with a zero-shot baseline. The fine-tuned models outperform the baseline on ROUGE-L, BLEU and BERTScore in every language and configuration. In a human preference study, low decoding temperatures receive the most preferences in all five languages and in both configurations, while the effect of the guidelines varies by language.
\end{abstract}

\section{Introduction}
Timely and effective communication is a critical aspect of crisis management: during an emergency, authorities must inform the population about what is happening and about what to do, and they must do it repeatedly as the situation evolves. This has to be done under time pressure and the volume of information handled during an event is a recognized source of stress for emergency operators \cite{misra2020information}. 
The issue is even worse in Europe, where nations are linguistically diverse (and many have multiple official languages) and there is a high cross-border mobility.
Therefore, warnings issued in a single language can fail to reach part of the population they are meant to protect. Issuing the same alert in several languages, at the speed an emergency requires, complicates the work of operators. Recent advances in Artificial Intelligence, and in particular Large Language Models (LLMs), which can interpret and produce text, offer a way to support this task \cite{foubert2026investigating}.

Crisis NLP has largely worked on analyzing text produced by users, with tasks such as detecting events in social media feeds and extracting informative content \cite{olteanu2014crisislex, alam2021crisisbench}, but recent work has also studied how to use LLMs to assist operators in writing warning messages for ongoing crises.
For instance, the \textit{CrisiText} dataset \cite{gonella-etal-2026-crisitext} contains crisis scenarios as chronological chains of events, each paired with a warning message.

\begin{figure*}
    \centering
    \includegraphics[width=1\linewidth]{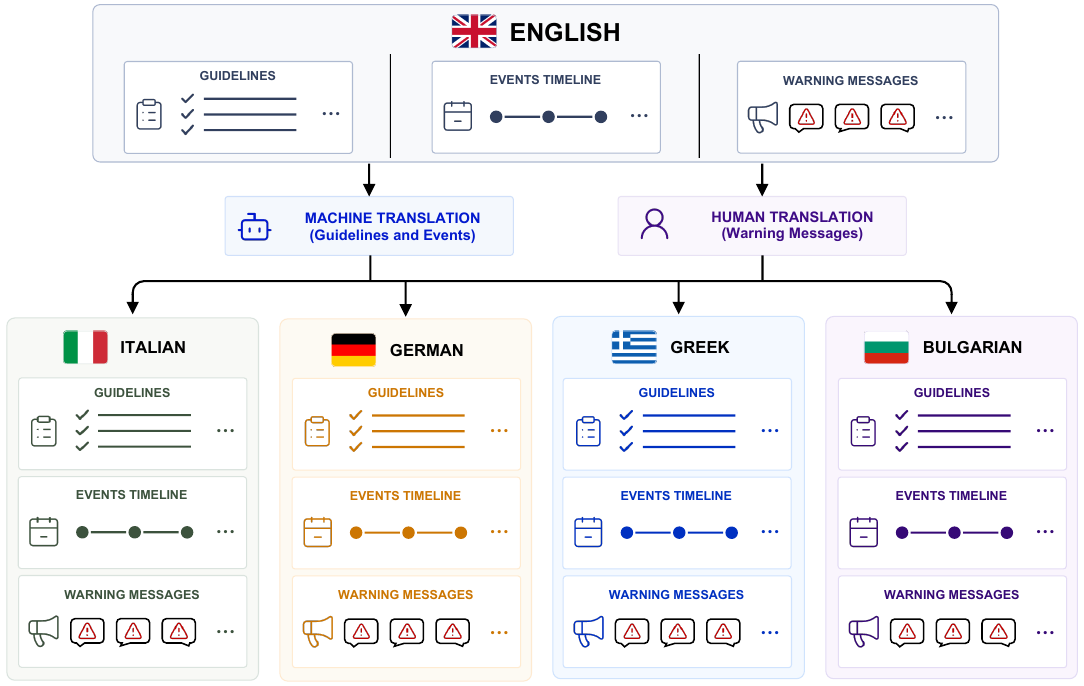}
    \caption{Workflow for the creation of \textit{CrisiLangs}. Event descriptions and FEMA guidelines are machine-translated into each target language, while the warning messages are translated and adapted by a native speaker.}
    \label{fig:dataset-structure}
\end{figure*}

Existing resources and related models such as \textit{CrisiText} focus only on English.
Multilingual coverage in crisis Natural Language Generation (NLG) has been addressed separately as a translation problem, by applying domain-adapted machine translation to existing crisis text \cite{staiano2025italert, castaldo2026translatingpressure}. Crisis content, however, is sensitive to automatic translation. 
\citet{ticona2026llm} report lower quality for low-resource languages and when translating out of English.
One way to avoid sending automatically translated content to civilians would be to generate messages directly in the target language. However, to the best of our knowledge, no comparable training resource is available for languages other than English.

To address this gap, we present \textit{CrisiLangs}\footnote{The dataset can be found at \url{https://huggingface.co/datasets/LanD-FBK/CrisiLangs}}, a multilingual extension of \textit{CrisiText} including Italian, Bulgarian, German, and Greek, chosen to cover different writing systems and alphabets.
In collaboration with crisis communication experts we manually translated a subset of the \textit{CrisiText} dataset into their native languages. The subsample covers 1,000 events from 411 crisis scenarios for a total of 5,000 event-message pairs across the five languages. 

We then use the resource to jointly fine-tune a multilingual model on all five languages in two configurations: with and without the emergency guidelines in the prompt. We tested the two configurations in two ways.
First, we evaluate the generations with automatic metrics at five decoding temperatures, with a zero-shot model used as the baseline. Since warning messages have to follow specific communication and behavioral guidelines, we varied the temperature to assess which level of lexical variation is best suited to the task. Second, to assess the quality of the generations, we asked native speakers to evaluate them in all five languages.

\section{Related Work}

Most NLP work on crises relies on social media data, and the resulting resources are designed for classification and information extraction rather than automatic generation \cite{imran2015processing, d-lewis-etal-2025-tapping}. Datasets of user-created content collected during floods, earthquakes or tornadoes \cite{olteanu2014crisislex, imran2016twitter} are annotated for informativeness, crisis detection or crisis type recognition, and are used to train classifiers \cite{alam2021crisisbench, liu2021crisisbert}. Fewer resources cover languages other than English: HumSet \cite{fekih-etal-2022-humset} provides humanitarian response documents in English, French and Spanish, and Kawarith \cite{alharbi-lee-2021-kawarith} collects Arabic crisis tweets. Relatedly, \citet{sanchez2023cross} study the cross-lingual transfer of classifiers trained on English crisis data to Spanish and Italian, and \citet{vitiugin2024multilingual} detect and rank help requests posted in English, Spanish and Turkish during natural disasters.

Crisis NLP does, however, include generation, mainly in the form of summarization. The CrisisFACTS track \cite{mccreadie2023crisisfacts} recasts disaster response as the production of fact-based daily summaries from social media and news, and recent systems address it by combining retrieval with instruction-following LLMs \cite{seeberger2024multi}. A related line works at the level of the individual emergency call, extracting the relevant information and presenting it to the call-taker \cite{otal2024llmassisted}. In both cases the output supports the response effort and is derived from text that already exists. Warning messages, on the other hand, are directly addressed to the population, and they are not produced by recapping existing textual reports.

How to reach speakers of different languages during an emergency has been studied in translation research under the name of crisis translation \cite{obrien2019crisistranslation}, which looks at language needs in disaster settings and at how translation can be included in emergency planning. In NLP, the same problem has been approached as machine translation. \citet{staiano2025italert} evaluate general-purpose LLM and NMT systems on Italian civil protection texts, while \citet{castaldo2026translatingpressure} fine-tune a small model for crisis-domain translation to produce simplified English output, proposed as a ``lingua franca'' when full multilingual coverage is not feasible. This transfer is not uniformly reliable: \citet{ticona2026llm} evaluate MT systems and LLMs on crisis-domain data in 30 languages and report substantial quality degradation, particularly for low-resource languages and in the direction from English into other languages.

Finally, a different line of work addresses the production of the message itself. The setting differs from the translation work described above, which starts from existing crisis text and moves it into another language. \textit{CrisiText} \cite{gonella-etal-2026-crisitext} pairs chronologically ordered crisis events with English warning messages written according to crisis guidelines, and has been used to fine-tune LLMs for message generation and post-editing. Other work addresses different communication settings: \citet{hong2025dynamic} generate replies to individual social media posts in which affected people express their needs. Guidance can also be grounded in the spatial layout of the affected area: \citet{gonella2026guideoutframeworkbenchmark} benchmark vision-language models on the generation of narrowcast guidance, i.e.\ guidance tailored to the individual civilian, in a simulated evacuation environment. These works produce messages in English, leaving the task in other languages unexplored.

\section{Multilingual Dataset}
\label{sec:data}
Training an LLM to produce warning messages requires event-message data pairs. In order to address the task in a multilingual setting, we built \textit{CrisiLangs} by extending the English \textit{CrisiText} data. 
Specifically, we used a manually curated subset of the original dataset and had its warning messages translated into Italian, Bulgarian, German, and Greek by four expert native speakers, one for each language.
The four languages were selected to ensure diversity in writing systems and language families, including three scripts
(Latin, Cyrillic and Greek) and four European language families (Romance, Slavic, Germanic and Hellenic). Figure~\ref{fig:dataset-structure} shows the workflow used to create the dataset.

\paragraph{Source Data.}

\textit{CrisiText} \cite{gonella-etal-2026-crisitext} contains more than 100,000 event-message pairs, where each event is labeled with one or more crisis categories (e.g., attack, flood or tornado). The messages were synthetically generated with an LLM following expert guidelines on communicative tone \cite{sutton2019alerts} and the behavioral guidance of the corresponding crisis categories, extracted from the official FEMA website.\footnote{\url{https://www.ready.gov/}} 
Events are also grouped into \textit{scenarios}, in the form of chronological chains of events describing the unfolding of a single crisis, and each scenario is entirely contained in one of the original training, development, and test splits.

\begin{figure*}
    \centering
    \includegraphics[width=0.98\linewidth]{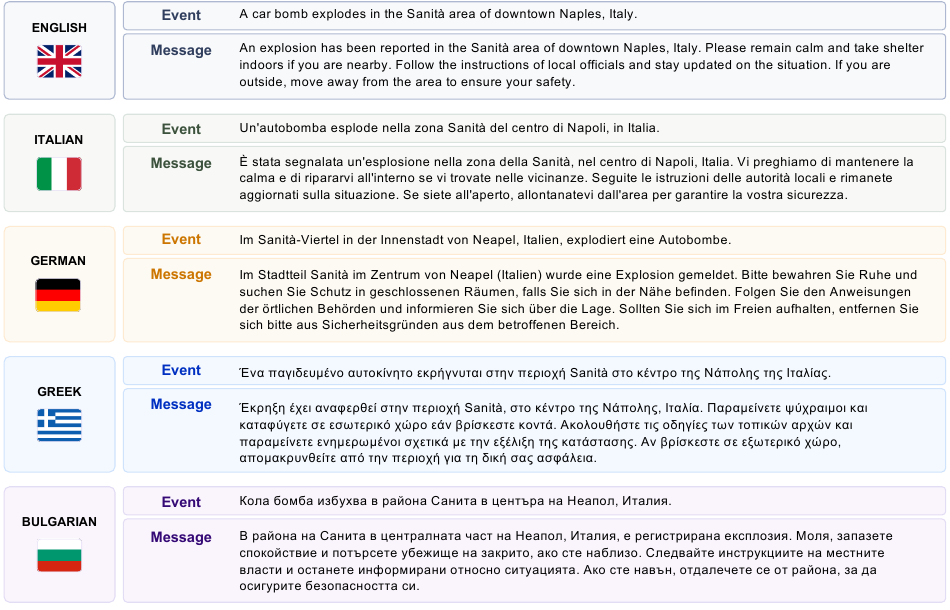}
    \caption{Example of event and warning message from \textit{CrisiLangs} in the five languages. The English event and message are the originals from \textit{CrisiText}. In the other four languages the messages are human translations produced by native speakers, while the events are machine-translated.}
    \label{fig:dataset-example}
\end{figure*}

\begin{table}[htbp]
\centering \small
\begin{tabular}{lrr}
\toprule
\textbf{Category} & \textbf{Scenarios} & \textbf{Events} \\
\midrule
\textbf{Human-caused} & \textbf{253} & \textbf{500} \\
\quad explosion & 105 & 201 \\
\quad fire & 76 & 153 \\
\quad attack & 72 & 146 \\
\midrule
\textbf{Natural} & \textbf{158} & \textbf{500} \\
\quad tornado & 33 & 111 \\
\quad hurricane & 26 & 95 \\
\quad flood & 21 & 59 \\
\quad winter weather & 14 & 49 \\
\quad heat & 17 & 48 \\
\quad thunderstorm & 16 & 48 \\
\quad avalanche & 10 & 35 \\
\quad earthquake & 7 & 12 \\
\quad landslide & 4 & 12 \\
\quad volcano & 3 & 12 \\
\quad wildfire & 4 & 10 \\
\quad tsunami & 3 & 9 \\
\midrule
\textbf{Total} & \textbf{411} & \textbf{1,000} \\
\bottomrule
\end{tabular}
\caption{Distribution of the event and message main categories over the 1,000 selected elements, grouped into human-caused and natural crises.}
\label{tab:multilingual-dataset-composition}
\end{table}

\paragraph{Selected Data.}

As part of the data quality assessment of \textit{CrisiText}, two crisis communication experts reviewed a sample of 1,000 event-message pairs, balanced between human-caused and natural disasters. 
We built our multilingual extension on this expert-reviewed sample. The pairs were drawn from the original training, development, and test splits in an 80/10/10 proportion.
Table~\ref{tab:multilingual-dataset-composition} reports the distribution of message and event categories. Events can carry multiple labels: the table reports only the main category of each event, i.e.\ the first label listed for its scenario, and omits the remaining labels assigned to the same scenario. The sample is balanced by construction, with 500 human-caused and 500 natural events.

\begin{table}[tbp]
\centering \small
\begin{tabular}{cccccc}
\toprule
\textbf{All} & \textbf{EN} & \textbf{IT} & \textbf{DE} & \textbf{EL} & \textbf{BG} \\
\midrule
306.9 & 268.6 & 316.4 & 324.1 & 331.1 & 294.4 \\
\bottomrule
\end{tabular}
\caption{Mean warning message length in characters per language in \textit{CrisiLangs}.}
\label{tab:multilingual-message-length}
\end{table}

\paragraph{Multilingual Translation.}

The subsample was given to four native speakers, one per target language (details about them are reported in Appendix~\ref{app:human-annotations}). Each translator received the original description of the event (both in English and automatically translated into their own language) and its associated warning message in English.
The FEMA guidelines used to generate the warning messages in \textit{CrisiText} were also automatically translated.\footnote{The translation was done with \texttt{GOOGLETRANSLATE} API.} Translators were asked to render the English message in their own language and to adapt it where necessary so that it would sound more natural and well formed.\footnote{Translators were allowed to use machine translation tools as support, provided that they revised the output.} In both the messages and the guidelines we replaced the US emergency number ``9-1-1'' with the European one, ``112'', since the target audience is European. This localized version of the guidelines is also the one used at training and generation time (Section~\ref{sec:experimental-design}).

Human translation was limited to the warning messages. This was a deliberate design choice: the message is what the model has to produce and what the human evaluation judges, whereas the event description and the guidelines only enter as input. Translation noise on the input side is therefore less critical, and it reflects a realistic deployment setting, where the input is processed automatically. Translators could rely on the English source in case of doubt.
In total, we obtained 5,000 event-message pairs (1,000 per language, including English), from 411 unique scenarios. Figure~\ref{fig:dataset-example} shows one such pair in all five languages, with the machine-translated event alongside the human-translated message. 
These parallel versions differ slightly in length, with per-language mean lengths reported in Table~\ref{tab:multilingual-message-length}.

\section{Experimental Design}
\label{sec:experimental-design}

We used the dataset to fine-tune a model in two configurations and to establish a zero-shot baseline for comparison. In order to assess which settings produce the most appropriate warning messages and how generation quality changes across the different languages, the generations were evaluated with automatic metrics and with a human preference study. The aim is to give future work on the resource a benchmark to compare against.

\paragraph{Model and Training Data Construction.}

Following the original \textit{CrisiText} work, we fine-tuned a single model, \texttt{Llama-3.1-8B-Instruct}, with LoRA \cite{hu2021loralowrankadaptationlarge}, keeping the same hyperparameters. To allow comparison with it we also kept the original prompt template (Figure~\ref{fig:prompt-example}), in which the instructions are in English while the events description are in the target language. In a preliminary zero-shot check we observed that the model consistently produced the warning message in the language of the input, so we did not add any explicit indication of the output language to the template. 
In particular, each training instance contains an event description to be used as the basis for the warning message, plus the preceding events description to be used as context. 
The message is generated only for the last event of the sequence.
Training was performed jointly across all five languages using a single multilingual model, rather than separate models for each language, with the aim of simplifying deployment in real-world settings.

\begin{figure}[ht]
\centering
\small
\setlength{\tabcolsep}{0pt}
\renewcommand{\arraystretch}{1.15}
\begin{tabular}{@{}p{0.97\columnwidth}@{}}
\toprule
\textsc{User} \\
\texttt{$^{\dagger}$\{Based on the provided guidelines,\}
Create a warning message informing on the current
happening, providing a suggestion, for the last line in
the following chain of events (be short, max 300
characters). No other output other than the message.}
\\[3pt]
$^{\dagger}$\texttt{Guidelines:}   \texttt{[GUIDELINES]} \\[3pt]
\texttt{Chain of events:} \texttt{[EVENTS]} \\
\bottomrule
\end{tabular}
\caption{Prompt example. The instruction is always in English, while \texttt{[EVENTS]} is replaced by the chain of events of the instance, in its language, with the target event as its last line.
The segments marked with $^{\dagger}$ appear only in the \textit{+guidelines} configuration, where \texttt{[GUIDELINES]} is replaced by the translated FEMA guidelines of the crisis categories of the event.}
\label{fig:prompt-example}
\end{figure}

\paragraph{Training Configurations.}
We trained the model in two configurations. In the \textit{plain} setup, the user prompt contains only the event descriptions. On the other hand, in the \textit{+guidelines} setup, the prompt additionally contains the translated FEMA guidelines of the crisis categories of the event. When an event belongs to more than one category, the corresponding guidelines are concatenated in the order in which the categories are listed. The comparison tests whether the guidelines have to be given explicitly at generation time, or whether fine-tuning on the data is enough for the model to infer them. Further details on the settings and on the hyperparameters are given in Appendix~\ref{app:training-setup}.

\begin{table*}[htbp]
\centering
\footnotesize
\setlength{\tabcolsep}{3pt}
\resizebox{\textwidth}{!}{%
\begin{tabular}{@{}llc*{5}{cccc}ccc}
\toprule
& & & \multicolumn{3}{c}{All} & & \multicolumn{3}{c}{English} & & \multicolumn{3}{c}{Italian} & & \multicolumn{3}{c}{German} & & \multicolumn{3}{c}{Greek} & & \multicolumn{3}{c}{Bulgarian} \\
\cmidrule(lr){4-6}\cmidrule(lr){8-10}\cmidrule(lr){12-14}\cmidrule(lr){16-18}\cmidrule(lr){20-22}\cmidrule(lr){24-26}
& & $T$ & R-L & B & BS & & R-L & B & BS & & R-L & B & BS & & R-L & B & BS & & R-L & B & BS & & R-L & B & BS \\
\midrule
\multirow{10}{*}{\rotatebox[origin=c]{90}{\textit{Zero-shot}}}
 & \multirow{5}{*}{\textit{plain}}
 & 1.0 & .168 & .037 & .878 & & .209 & .072 & .890 & & .175 & .041 & .882 & & .185 & .038 & .884 & & .130 & .019 & .865 & & .143 & .016 & .871 \\
 & & 0.8 & .184 & .043 & .884 & & .232 & .083 & .893 & & .196 & .045 & .886 & & .202 & .047 & .888 & & .145 & .021 & .874 & & .148 & .021 & .878 \\
 & & 0.6 & .201 & .048 & .888 & & .234 & .084 & \textbf{.895} & & .215 & .049 & .891 & & .218 & .055 & .892 & & .176 & .032 & .883 & & .166 & .021 & .883 \\
 & & 0.4 & .205 & .051 & .890 & & .232 & .083 & .894 & & .214 & .053 & .890 & & .226 & .050 & .894 & & .174 & .036 & .887 & & .181 & .035 & .887 \\
 & & 0.2 & \textbf{.211} & \textbf{.056} & \textbf{.892} & & \textbf{.240} & \textbf{.090} & .894 & & \textbf{.221} & \textbf{.056} & \textbf{.892} & & \textbf{.233} & \textbf{.059} & \textbf{.896} & & \textbf{.182} & \textbf{.039} & \textbf{.888} & & \textbf{.182} & \textbf{.035} & \textbf{.889} \\
\cmidrule(lr){2-26}
 & \multirow{5}{*}{\textit{+guid.}}
 & 1.0 & .187 & .049 & .881 & & .231 & .080 & .892 & & .175 & .047 & .881 & & .211 & .056 & .884 & & .146 & .032 & .872 & & .170 & .031 & .878 \\
 & & 0.8 & .196 & .054 & .885 & & .226 & .068 & .890 & & .191 & .056 & .886 & & .236 & .072 & .892 & & .154 & .040 & .878 & & .171 & .031 & .882 \\
 & & 0.6 & .207 & .056 & .889 & & .247 & .088 & .895 & & .196 & .051 & .888 & & .229 & .066 & .890 & & .174 & .035 & .883 & & .191 & .038 & .888 \\
 & & 0.4 & .218 & .067 & .890 & & .248 & .094 & .894 & & .213 & .067 & .890 & & .247 & .076 & .893 & & .178 & \textbf{.048} & \textbf{.885} & & \textbf{.202} & \textbf{.051} & \textbf{.890} \\
 & & 0.2 & \textbf{.220} & \textbf{.069} & \textbf{.891} & & \textbf{.259} & \textbf{.104} & \textbf{.896} & & \textbf{.216} & \textbf{.070} & \textbf{.891} & & \textbf{.247} & \textbf{.077} & \textbf{.894} & & \textbf{.179} & .044 & .884 & & .198 & .050 & \textbf{.890} \\
\midrule
\multirow{10}{*}{\rotatebox[origin=c]{90}{\textit{Fine-tuned}}}
 & \multirow{5}{*}{\textit{plain}}
 & 1.0 & .259 & .111 & .902 & & .311 & .159 & .915 & & .245 & .106 & .901 & & .278 & .110 & .903 & & .236 & .089 & .893 & & .224 & .079 & .897 \\
 & & 0.8 & .284 & .134 & .907 & & .350 & .189 & .920 & & .269 & .125 & .906 & & .308 & .137 & .909 & & .256 & .111 & .900 & & .235 & .093 & .902 \\
 & & 0.6 & .307 & .158 & .912 & & .366 & .211 & .922 & & .297 & .145 & .912 & & .328 & .161 & .913 & & .282 & .138 & .905 & & .258 & .122 & .908 \\
 & & 0.4 & .320 & .169 & .914 & & .387 & .226 & .924 & & .308 & .159 & .912 & & .340 & .176 & .915 & & .291 & .136 & .910 & & .271 & .123 & .910 \\
 & & 0.2 & \textbf{.331} & \textbf{.177} & \textbf{.916} & & \textbf{.389} & \textbf{.227} & \textbf{.925} & & \textbf{.331} & \textbf{.171} & \textbf{.915} & & \textbf{.345} & \textbf{.177} & \textbf{.916} & & \textbf{.306} & \textbf{.150} & \textbf{.911} & & \textbf{.281} & \textbf{.138} & \textbf{.912} \\
\cmidrule(lr){2-26}
 & \multirow{5}{*}{\textit{+guid.}}
 & 1.0 & .258 & .116 & .901 & & .342 & .190 & .917 & & .240 & .101 & .901 & & .269 & .112 & .900 & & .215 & .081 & .892 & & .220 & .082 & .896 \\
 & & 0.8 & .287 & .135 & .907 & & .370 & .214 & .921 & & .258 & .114 & .905 & & .303 & .132 & .907 & & .260 & .121 & .901 & & .243 & .089 & .902 \\
 & & 0.6 & .308 & .154 & .911 & & .378 & .221 & .922 & & .299 & .145 & .910 & & .329 & .163 & .913 & & .286 & .136 & .907 & & .247 & .099 & .905 \\
 & & 0.4 & .322 & .165 & \textbf{.914} & & \textbf{.396} & \textbf{.233} & \textbf{.924} & & .297 & .149 & .911 & & \textbf{.341} & \textbf{.172} & \textbf{.915} & & .299 & .144 & .909 & & .277 & .120 & .910 \\
 & & 0.2 & \textbf{.330} & \textbf{.171} & \textbf{.914} & & .393 & .232 & .922 & & \textbf{.319} & \textbf{.162} & \textbf{.913} & & \textbf{.341} & .167 & .914 & & \textbf{.318} & \textbf{.159} & \textbf{.910} & & \textbf{.283} & \textbf{.128} & \textbf{.911} \\
\bottomrule
\end{tabular}}
\caption{ROUGE-L F1 (R-L), BLEU (B) and BERTScore F1 (BS) on the test split, per configuration, temperature and language. Bold marks the best temperature within each block. Zero-shot uses the un-finetuned \texttt{Llama-3.1-8B-Instruct} with identical configuration.}
\label{tab:auto-metrics}
\end{table*}

\paragraph{Generation and Automatic Metrics.}
After training, we generated warning messages for the 100 test events in all five languages, in both configurations and at five decoding temperatures (1.0, 0.8, 0.6, 0.4 and 0.2), to assess the trade-off between output variety and conformity to the crisis guidelines. As a baseline, we also ran the same generations with the off-the-shelf model in a zero-shot setting, using the same prompt as in training. 
In order to measure the effect of fine-tuning and of the decoding temperature, all generations were evaluated against the reference messages with two overlap metrics, ROUGE-L and BLEU, and with BERTScore, which is a semantic metric based on contextual embeddings. These scores were then used to select the data included in the human study.
More information on the metrics is given in Appendix~\ref{app:automatic-metrics}.

\paragraph{Human Preference Study.}
Since automatic metrics are only a proxy for the quality of a warning message, we complemented them with a human preference study.
On the basis of the automatic evaluation we restricted the study to the three best-performing decoding temperatures.
For each of the two training configurations, one native speaker per language was presented with 50 test event-message pairs in their own language, each accompanied by the three corresponding generations. The two configurations were annotated separately, and the three candidates were shown anonymized, without any indication of the temperature that distinguished them. Annotators were asked to choose the message they considered grammatically better and more compliant with the task, and could leave free-text comments to motivate the choice or report problems the selection alone could not express. At the end of the annotation, we also asked which of the two configurations they preferred overall.

\section{Results}
\label{sec:results}

This section reports what the experiments described in Section~\ref{sec:experimental-design} indicate about the resource. The automatic evaluation serves two purposes: it establishes whether fine-tuning on the resource is necessary, and it narrows the set of decoding temperatures presented to the annotators. The human preference study then addresses what the metrics cannot capture, namely whether the generations are appropriate in each language and which of the two training configurations should be preferred.

\subsection{Automatic Evaluation}
\label{sec:results-automatic}

Table~\ref{tab:auto-metrics} reports ROUGE-L, BLEU and BERTScore on the test split for each language and decoding temperature, in the two training configurations and in the corresponding baselines. The comparison is intended as intra-language, across temperatures and configurations. All three metrics are sensitive to language-specific factors such as tokenization and morphology, so inter-language differences do not indicate a quality ranking.

The fine-tuned models score higher than the baseline on all three metrics in every language, temperature and prompt configuration.
ROUGE-L and BLEU reward two aspects of a generation at once, its length and its wording. BLEU keeps them separate, as the product of the brevity penalty and the n-gram precision term.
As Table~\ref{tab:length-metrics-02} shows, at $T=0.2$ the fine-tuned models match the length of the human references, while the baseline with the plain prompt produces little more than half of the 300 characters. The table also reports BLEU without the brevity penalty: the n-gram precision term rises from .114 to .177, so fine-tuning also improves the wording.
Adding the guidelines to the baseline prompt recovers part of the length gap, but the gain is small compared to fine-tuning and the n-gram precision term does not improve (.114 to .104).
BERTScore shows the same pattern: at $T{=}0.2$ fine-tuning raises it from .892 to .916, while adding the guidelines to the baseline prompt leaves it virtually unchanged (.891).
Therefore, these results indicate that fine-tuning on this data contributes substantially to both the length and the wording of the generations, and we restrict the rest of the evaluation to the fine-tuned models.

\begin{table}[htbp]
\centering
\small
\begin{tabular}{lcc}
\toprule
 & Length & BLEU w/o BP \\
\midrule
Gold & 304 & -- \\
\midrule
ZS, plain & 173 & .114 \\
ZS, +guid. & 205 & .104 \\
FT, plain & 295 & .177 \\
FT, +guid. & 299 & .171 \\
\bottomrule
\end{tabular}
\caption{Mean generation length in characters and BLEU without the brevity
penalty (BP), at $T{=}0.2$, over all five languages, for the zero-shot (ZS)
and fine-tuned (FT) models. Per-language values are in Appendix~\ref{app:automatic-metrics}.}
\label{tab:length-metrics-02}
\end{table}

Across all setups, zero-shot and fine-tuned alike, scores rise as temperature decreases. 
This pattern reflects the fact that lower temperatures sharpen the probability around the most likely sequence, yielding wording that is less likely to include rare or unexpected choices, and thus closer to the reference text (which these metrics reward by construction). Whether such messages are also better is assessed in the human study (Section~\ref{sec:results-human}).
When results are pooled across languages, both ROUGE-L and BLEU increase in a strictly monotonic way from temperature 1.0 down to 0.2. BERTScore does the same, except for a tie between 0.4 and 0.2 in the fine-tuned \mbox{+guidelines} configuration (.914). At the language level, however, the trend holds but is not always monotonic: in the \mbox{+guidelines} configuration, for example, ROUGE-L for English and BLEU for German are marginally lower at 0.2 than at 0.4. In fine-tuned models these deviations are less relevant, never exceeding .005.

The two training configurations are close to each other. In the 0.6-0.2 range, ROUGE-L and BLEU favor the \mbox{+guidelines} configuration for English and Greek, while the plain configuration is better on Italian. On German the two are close, with the plain configuration slightly ahead at temperature 0.2. Bulgarian is the only language where the two overlap metrics disagree: at the two lowest temperatures the plain configuration is ahead on BLEU and the \mbox{+guidelines} one on ROUGE-L. BERTScore separates the two configurations even less, with differences of at most .003 in every language in this range, and does not reproduce the advantage of the guidelines on English and Greek. Considering the aggregated metrics over all languages, the two configurations are virtually identical. No configuration is consistently better, so the automatic metrics do not settle which configuration to prefer, motivating the human evaluation described in Section~\ref{sec:results-human}.

Based on the automatic scores, we focused on temperatures 0.6, 0.4 and 0.2 for the human evaluation.

\subsection{Human Preference Study}
\label{sec:results-human}

The automatic scores rank the temperatures consistently, but they do not indicate which generation a reader would prefer, and they do not separate the two training configurations.
We therefore asked one native speaker per language, with expertise in crisis communication, to compare the three generations and to select the one they considered most natural, grammatical and consistent with the task.
The annotation covered 50 test events, with the candidates generated at temperatures 0.6, 0.4 and 0.2, and was repeated for both training configurations. When no candidate was acceptable, the annotator could leave the item unannotated.

\begin{table}[t]
\centering\small
\begin{tabular}{lrrrrrrrr}
\toprule
& \multicolumn{4}{c}{\textit{plain}} & \multicolumn{4}{c}{\textit{+guidelines}} \\
\cmidrule(lr){2-5}\cmidrule(lr){6-9}
& 0.6 & 0.4 & 0.2 & $N$ & 0.6 & 0.4 & 0.2 & $N$ \\
\midrule
English   &  8 & 15 & 27 & 50 &  12 & 8 & 30 & 50 \\
Italian   &  6 & 15 & 29 & 50 &  7 & 15 & 27 & 49 \\
German    & 12 & 13 & 24 & 49 &  7 & 17 & 26 & 50 \\
Greek     & 11 &  2 & 37 & 50 &  4 & 22 & 24 & 50 \\
Bulgarian &  4 & 17 & 29 & 50 &  3 &  9 & 25 & 37 \\
\bottomrule
\end{tabular}
\caption{Preference counts per configuration and language. $N$ is the number of annotated events out of 50.}
\label{tab:preferences}
\end{table}

\paragraph{Temperature.}

Table~\ref{tab:preferences} shows that temperature 0.2 receives the most preferences in all five languages and in both training configurations, although the trend is not uniformly evident in all cases (e.g., for the \mbox{+guidelines} model).
In Bulgarian, the generations using the guidelines led to the rejection of all three candidates in 13 items because of errors in the generated text.
In the plain configuration a choice was always made, although the comments indicate that some candidates were still flawed. Lastly, Greek in the \mbox{+guidelines} configuration shows the smallest margin in the table, with only two preferences separating temperature 0.2 and 0.4, so the annotation does not indicate a clear preference between them.

\paragraph{Training Configuration.}

At the end of the task, each annotator also indicated which of the two models they preferred overall. All chose the plain model except the Greek annotator, who preferred the \mbox{+guidelines} one. The Italian annotator reported the plain model to have a better use of terminology and more specific warnings about the event, and the German annotator a better quality of the German text and slightly more accurate warning messages. Similarly, the English annotator motivated the preference with the accuracy of the messages, although reporting a narrower difference between the two models. The Bulgarian annotator's justification was negative, referring to the amount of non-compliant content produced by the \mbox{+guidelines} model. On the other hand, the Greek annotator described the \mbox{+guidelines} generations as clearer, better structured and more effective in conveying urgency.
ROUGE-L and BLEU favor \mbox{+guidelines} in the 0.6-0.2 range only for Greek and English, matching the Greek annotator but not the English one, while BERTScore does not separate the two configurations in either language.

\paragraph{Qualitative Feedback.}

In addition to the preference judgment, annotators could leave free-text comments. These were optional and used to different extents, so they are reported as qualitative evidence rather than as a systematic annotation. They point to a difference between the two configurations that the automatic metrics do not capture.
The Greek, Italian and English annotators motivated their choices based on the wording of the warning and the appropriateness of the recommended actions. 
In Greek these reasons differ between the two configurations: the plain model was chosen for its simpler wording and calmer tone, the \mbox{+guidelines} model for the specificity of its information and of the recommended actions.
In Bulgarian and German the comments instead report defects in the generations, most frequently grammatical errors and unfinished text. These are concentrated on the \mbox{+guidelines} model, consistent with the overall preference for the plain configuration reported above. None of these problems is visible in the automatic scores, and this is where the human study contributes most.

\section{Discussion}
\label{sec:discussion}

Our experiments show that fine-tuning on \textit{CrisiLangs} yields a single model that produces warning messages in all five languages, above the baseline in the automatic evaluation, but the human study suggests that quality is not uniform across languages. For English, Italian, German and Greek, the annotators expressed a preference for almost every item, and their comments mostly point to specific flaws rather than to unusable generations.
By contrast, with the Bulgarian plain model a choice could always be made, even when candidates were non-compliant or contained errors. With the \mbox{+guidelines} model, 13 items were left unannotated because none of the three candidates was acceptable.
The resource therefore supports warning generation in all five languages, with the Bulgarian \mbox{+guidelines} output being less reliable, mostly for limitations of the LLM.

Across the five languages we evaluated, low decoding temperatures are consistently preferred. The automatic metrics increase as the temperature decreases, as they favor wording closer to the reference by construction, so the more informative evidence comes from the human study: among the three temperatures compared, 0.2 receives the most preferences in all languages and in both training configurations (Table~\ref{tab:preferences}). Since annotators were asked to judge grammaticality and compliance with the task, this indicates that they found lower-temperature messages better on these criteria. The pattern holds in both the plain and the \mbox{+guidelines} configuration, so it does not depend on having the guidelines in the prompt. For practical use of this resource, low temperatures appear to be the safer choice.

The effect of the guidelines in the prompt is less uniform. Four of the five annotators preferred the plain model overall, citing better wording and more relevant content, and only the Greek annotator preferred the \mbox{+guidelines} one, for its clarity and structure. 
These preferences, together with the automatic metrics (where neither configuration is consistently ahead), suggest that in-prompt guidelines are not consistently beneficial. One possible explanation is that fine-tuning on the gold messages already induces much of the intended compliance, making the guidelines largely redundant; moreover, adding a long block of instructions increases prompt length and can dilute attention to the event-specific details, which may ultimately hurt generation quality.

A final observation concerns the relation between the two evaluations. The automatic metrics rank the temperatures consistently, but none of them, BERTScore included, anticipates the Bulgarian result: in Table~\ref{tab:auto-metrics} the \mbox{+guidelines} scores for Bulgarian are close to the plain ones at every temperature, yet the annotator found none of the three candidates acceptable in 13 of the 50 items in that configuration and in none of the plain ones. Human validation therefore remains necessary when crisis warning generation is extended to new languages.

\section{Conclusion}

We presented \textit{CrisiLangs}, a multilingual extension of \textit{CrisiText} to Italian, Bulgarian, German, and Greek.
The resource contains 1,000 events from 411 scenarios, with human-translated warning messages in four languages alongside the original English ones, for a total of 5,000 event-message pairs.

To benchmark this resource we fine-tuned \texttt{Llama-3.1-8B-Instruct} on the five languages and compared it with a zero-shot baseline.
We evaluated the generations in the five languages both with automatic metrics and with a human preference study.
We observed that, compared to the baseline, fine-tuning improved ROUGE-L, BLEU and BERTScore in every language and prompt configuration (with or without guidelines).
Lower decoding temperatures lead to higher automatic scores, which can be expected from reference-based metrics, this is then confirmed by the annotators where the temperature 0.2 received the most preferences in all five languages and in both configurations.

The human evaluation suggests that providing the emergency guidelines to the model during generation does not reliably improve the output: four of the five annotators preferred the plain configuration overall, and only the Greek annotator preferred the configuration with guidelines.
In our evaluation, the automatic metrics did not reveal the problems found by the annotators: the Bulgarian \mbox{+guidelines} generations score close to the plain ones, yet in 13 of the 50 items none of the candidates was acceptable. This points to the importance of expert native-speaker assessment when benchmarking warning message generation.

\section*{Limitations}
\label{sec:limitations}
In the current setup the model receives the event in the target language and generates the warning message in the same language, so automatic translation is used only on the input side. The resource also supports feeding the model the English event directly and generating the message in the target language in a single pass, removing automatic translation from the pipeline entirely, but we do not evaluate this setting, as it would require a further round of automatic and human evaluation.

A second limitation is that we focus on fine-tuning a single multilingual model. Complementary comparisons (e.g., comparisons with language-specific pretrained LLMs, particularly for Greek and Bulgarian, which use non-Latin scripts, and with a model fine-tuned exclusively on English data to assess the added value of target-language supervision) are natural next steps that we leave to future work.
Translation and preference annotation were carried out by a single native speaker per language, reflecting the specialized expertise required for crisis communication. Future work with additional annotators would enable measuring inter-annotator agreement; nonetheless, the preference for temperature 0.2 is shared by all five annotators.

Finally, the resource provides 1,000 expert-reviewed event-message pairs per language; by comparison, \textit{CrisiText} contains more than 100,000 English pairs, as our extension focuses on the expert-reviewed subset.
Human translation was also limited to the warning messages: the messages are what the model produces and what the annotators judge, whereas the events and guidelines only enter as input. Translation noise on the input side therefore remains in the data.

\bibliography{custom}

\clearpage

\appendix

\section{Hyperparameters and Generation Settings}
\label{app:training-setup}

\paragraph{LoRA Adapter Configuration.}
We fine-tuned \texttt{meta-llama/Llama-3.1-8B-Instruct} with a \texttt{peft} LoRA adapter: rank $16$, alpha $16$, dropout $0$, no bias adaptation, task type \texttt{CAUSAL\_LM}. The adapter targets \texttt{q\_proj}, \texttt{k\_proj}, \texttt{v\_proj}, \texttt{o\_proj}, \texttt{gate\_proj}, \texttt{up\_proj} and \texttt{down\_proj}. The base model was loaded in \texttt{bfloat16} without quantization, and the adapter was merged back into the base weights before generation.

\paragraph{SFT.}
Training used \texttt{trl}'s \texttt{SFTTrainer}. Sequences were capped at $1024$ tokens; the learning rate was $3\mathrm{e}{-4}$ with a cosine schedule, weight decay $0.01$ and warmup ratio $0.03$. We used a batch size of $2$ for training and $4$ for evaluation, with gradients accumulated over $4$ steps, for an effective batch size of $8$. We ran for up to $10$ epochs, evaluating and checkpointing every $50$ steps with early stopping patience $5$, and kept the checkpoint with the lowest validation loss. Data were shuffled with seed $42$ and the trainer ran with seed $0$. The two configurations differ only in the prompt: \textsc{plain} receives the instruction and the chain of events, \textsc{+guidelines} additionally receives the localized guideline blocks. Both were trained with the same hyperparameters on the same $4{,}000$ training examples.

\paragraph{Chat Formatting.}
Padding used Llama-3.1's \texttt{<|finetune\_right\_pad\_id|>} on the right, and turns were formatted with the Llama-3.1 chat template.

\paragraph{Llama Generation.}
Decoding used sampling with a top-p of $1$ and a cap of $128$ new tokens, sufficient for a warning message. Each test prompt was decoded once per temperature ($1$, $0.8$, $0.6$, $0.4$, $0.2$), giving $2{,}500$ generations per model configuration ($100$ test records $\times$ $5$ languages $\times$ $5$ temperatures).

\paragraph{Zero-shot Baseline.}
The zero-shot baseline was the same \texttt{meta-llama/Llama-3.1-8B-Instruct} without the LoRA adapter, and was run over the same test split with the same prompts, the same five temperatures and the same decoding arguments and post-processing.

\paragraph{Compute.}
All fine-tuning and generation runs were performed on a single NVIDIA A40 GPU.

\section{Automatic Metrics}
\label{app:automatic-metrics}

\paragraph{Evaluation Packages.}
For automatic metrics we used Hugging Face's \texttt{evaluate} library (\texttt{evaluate} $0.4.6$, \texttt{rouge-score} $0.1.2$, \texttt{nltk} $3.10.2$) for ROUGE, \texttt{sacrebleu} $2.6.0$ for BLEU, and \texttt{bert-score} $0.3.13$ (\texttt{transformers} $4.57.6$) for BERTScore, all computed per language, configuration and temperature. 
\paragraph{ROUGE-L} is the F1 of the longest common subsequence, computed over the message as a single token sequence, and aggregated as the median of $1{,}000$ bootstrap resamples. \texttt{rouge\_score}'s default tokenizer lowercases and keeps only Latin letters and digits, discarding all Cyrillic and Greek characters: with this configuration only about $2\%$ of Bulgarian and $5\%$ of Greek reference tokens survive, and what remains is digits and Latin-script names. We therefore passed a Unicode-aware tokenizer, so that all five languages are scored on their own scripts. 
\paragraph{BLEU} was computed at corpus level with SacreBLEU's default settings\footnote{Signature: \nolinkurl{nrefs:1|case:mixed|eff:no|tok:13a|smooth:exp|version:2.6.0}}, including the script-agnostic \texttt{13a} tokenizer. Scores were divided by $100$ to match the $0$--$1$ scale of the other metrics.
\paragraph{BERTScore} was computed with \texttt{xlm-roberta-large} for all five languages, without idf weighting and without baseline rescaling, as no rescaling baseline is available for Bulgarian and Greek. Each reported score is the mean of the per-item F1 scores. Without rescaling, all scores in Table~\ref{tab:auto-metrics} fall between .865 and .925, so BERTScore differences are small in absolute terms and should only be compared within its own columns.

All reported scores were computed on CPU directly from the released generation files.

\paragraph{Per-Language Metrics.}
Table~\ref{tab:length-metrics-full} reports the per-language values behind the aggregated numbers of Table~\ref{tab:length-metrics-02}. Both components move in the same direction in all five languages: the fine-tuned models are closer to the reference length than the corresponding baseline, and the n-gram precision term is higher in every language and in both prompt configurations.

\begin{table}[htbp]
\centering
\small
\begin{tabular}{lcccccc}
\toprule
 & All & EN & IT & BG & DE & EL \\
\midrule
\multicolumn{7}{l}{\textit{mean length (characters)}} \\
Gold & 304 & 267 & 315 & 289 & 317 & 330 \\
ZS, plain & 173 & 188 & 163 & 163 & 182 & 168 \\
ZS, +guid. & 205 & 204 & 214 & 192 & 221 & 195 \\
FT, plain & 295 & 269 & 297 & 272 & 326 & 309 \\
FT, +guid. & 299 & 270 & 312 & 284 & 317 & 314 \\
\midrule
\multicolumn{7}{l}{\textit{BLEU without brevity penalty}} \\
ZS, plain & .114 & .137 & .129 & .076 & .117 & .101 \\
ZS, +guid. & .104 & .139 & .102 & .080 & .112 & .080 \\
FT, plain & .177 & .227 & .184 & .138 & .177 & .156 \\
FT, +guid. & .171 & .232 & .162 & .128 & .167 & .162 \\
\bottomrule
\end{tabular}
\caption{Per-language mean generation length in characters and BLEU without the brevity penalty, at $T{=}0.2$, for the zero-shot (ZS) and fine-tuned (FT) models with the plain and +guidelines (+guid.) prompts. As in Table~\ref{tab:auto-metrics}, the comparison is intra-language. Languages are English (EN), Italian (IT), Bulgarian (BG), German (DE) and Greek (EL).}
\label{tab:length-metrics-full}
\end{table}

\section{Human Annotations}
\label{app:human-annotations}

The annotators were members of organizations engaged in crisis management, selected if native speakers in the target language and according to their professional experience. For the translation and annotation work, annotators were compensated fairly by their respective organizations in compliance with applicable national laws.

For each language and each of the two model configurations, annotators compared the outputs of the three best-scoring temperatures ($0.6$, $0.4$, $0.2$) on $50$ test events. The three outputs were presented as blind candidate A/B/C, and annotators selected one candidate per event and could optionally leave a free-text comment.

Figure~\ref{fig:annotator-guidelines} reproduces verbatim the instructions given to the annotators for the translation and for the preference task.

\begin{figure}[ht]
\centering
\small
\setlength{\tabcolsep}{0pt}
\renewcommand{\arraystretch}{1.15}
\begin{tabular}{@{}p{0.97\columnwidth}@{}}
\toprule
\textsc{Translation task} \\[2pt]
The task consists in creating alert messages in your
language based on the English alert messages we provide.
We ask you to translate the messages and, when necessary,
adapt them to sound more natural and appropriate as alert
messages in your language. \\
\midrule
\textsc{Preference task} \\[2pt]
Your task is to: \\[1pt]
-- For each row, select your preferred message using the
dropdown menu, based on grammaticality and consistency
with the task. \\[1pt]
-- At the top of each sheet, indicate which configuration
(column) you preferred overall, and briefly explain why. \\[1pt]
-- Let us know which of the two models (sheets) you
preferred overall, and why. \\[3pt]
(Note: there is no need to correct the text of the
messages.) \\
\bottomrule
\end{tabular}
\caption{Instructions given to the annotators, reproduced
verbatim. The translation instructions were provided together
with the English source; the preference instructions accompanied
one spreadsheet per training configuration.}
\label{fig:annotator-guidelines}
\end{figure}

\end{document}